\documentclass{article}
\usepackage[preprint]{NIPS}

\usepackage[utf8]{inputenc} 
\usepackage[T1]{fontenc}    
\usepackage{url}            
\usepackage{booktabs}       
\usepackage{amsfonts}       
\usepackage{nicefrac}       
\usepackage{microtype}      

\usepackage{graphicx}
\usepackage{multirow}
\usepackage{subfigure}
\usepackage{amsmath}
\usepackage{hyperref}
\hypersetup{
    colorlinks=true,
    linkcolor=black
}

\usepackage[misc]{ifsym}

\title{A Multi-Scale Framework for Out-of-Distribution Detection in Dermoscopic Images}

\author{%
  Zhongzheng Huang \\
  Fuzhou University\\
  \texttt{200327039@fzu.edu.cn} \\
  \And
  Tao Wang\thanks{Corresponding author. Paper accepted by the 4th International Conference on Machine Learning for Cyber Security (ML4CS 2022), Guangzhou, China. The final authenticated version is
available online at \url{https://doi.org/10.1007/978-3-031-20096-0_12}.} \\
  Minjiang University\\
  \texttt{twang@mju.edu.cn} \\
  \And
  Yuanzheng Cai\\
  Minjiang University\\
  \texttt{yuanzheng\_cai@mju.edu.cn} \\
  \And
  Lingyu Liang \\
  South China University of Technology\\
  \texttt{eelyliang@scut.edu.cn} \\
}

\begin{document}

\maketitle

\begin{abstract}
The automatic detection of skin diseases via dermoscopic images can improve the efficiency in diagnosis and help doctors make more accurate judgments. However, conventional skin disease recognition systems may produce high confidence for out-of-distribution (OOD) data, which may become a major security vulnerability in practical applications. In this paper, we propose a multi-scale detection framework to detect out-of-distribution skin disease image data to ensure the robustness of the system. Our framework extracts features from different layers of the neural network. In the early layers, rectified activation is used to make the output features closer to the well-behaved distribution, and then an one-class SVM is trained to detect OOD data; in the penultimate layer, an adapted Gram matrix is used to calculate the features after rectified activation, and finally the layer with the best performance is chosen to compute a normality score. Experiments show that the proposed framework achieves superior performance when compared with other state-of-the-art methods in the task of skin disease recognition.
\end{abstract}
\section{Introduction}
Skin is the largest organ of the human body, and skin diseases contributed $1.79\%$ to the global burden of disease measured in disability-adjusted
life years~\cite{karimkhani2017global}. Therefore, computer-aided monitoring, diagnosis and management of skin diseases are of wide interest to the medical imaging community. In particular, deep learning has been recently used in this field for the task of automatic diagnosis of skin diseases~\cite{esteva2017dermatologist}. Typically, deep neural networks assume that the training set and the test set are of the same set of classes~\cite{goodfellow2016deep}. However, in real-life applications, the recognition system often needs to detect some images that do not belong to the training classes~\cite{liang2017enhancing}. In this case, conventional systems may generate high confidence in some images without skin diseases. In addition, it is important for these systems to identify irrelevant images (e.g., animal pictures with color similar to that of skin lesions). Failure to do so weakens the security of a skin disease identification system and adversely affects its use in general.

In order to address the above problem, out-of-distribution~(OOD) detection methods in deep learning are proposed to reduce the error rate of the model by identifying in advance whether the input image is an OOD sample~\cite{hendrycks2016baseline}. In broad terms, existing OOD detection methods can be categorized into density-based, distance-based and classification-based~\cite{yang2021generalized}. In particular, classification-based OOD detection methods judge whether a sample is OOD by using a classifier to classify the extracted features~\cite{liu2020energy}. Inspired by the success of multi-scale detection models in object detection \cite{zhang2021empirical}, we design a classification-based multi-scale detection framework to further improve OOD detection performance. First, we use different classifiers for the features extracted from different layers of the network. In the shallow layers, we choose one-class SVM for classification, which does not require a large amount of data to train a good classifier and hence fits the task of skin disease detection with a relatively small amount of data; in the penultimate layer, we choose an adapted Gram matrix that calculates the correlation between the ID features and the OOD features, and finally obtain a relatively accurate detection score. In addition, in order to obtain features that are closer to the well-behaved distribution for computation, we add a rectified activation operation after feature extraction for each layer of the model, which selects the final feature by comparing it with a preset threshold and sending it to the classifier corresponding to the current layer. Our main contributions are as follows:

$\cdot$ We propose a multi-scale detection framework that integrates one-class SVM and adapted Gram matrix to detect and compare features at different layers of the deep neural network, and then selects the layer with the best performance to compute the final normality score.

$\cdot$ We introduce a rectified activation operation after each deep neural network layer to produce a well-behaved distribution for the subsequent feature classifiers.

$\cdot$ We compare our method with recently proposed OOD detection methods on multiple datasets and models, and the results show that our method is able to achieve the state-of-the-art in most settings.

\section{Related Work}

\subsection{Out-of-Distribution Detection}
In recent years, out-of-distribution detection has been widely studied in the field of image classification. For example, Zaeemzadeh et al.~\cite{zaeemzadeh2021out} show that embedding in-distribution~(ID) data into a low-dimensional space can make OOD data easier to detect. If the probability of the test data occupying an area with ID data is $0$, it belongs to OOD. Zisselman et al.~\cite{zisselman2020deep} introduce a method of learning residual distribution from base Gaussian distribution for building flow structures. Serr{\`a} et al. \cite{serra2019input} observe that generative models are ineffective for OOD detection, and they use an estimate of input complexity to obtain OOD scores. Yang et al. \cite{yang2021semantically} propose a semantically coherent OOD detection benchmark, and design a framework for extracting features with unsupervised dual grouping, which enriches semantic information while improving the classification of ID data and the detection of OOD data.

\subsection{Out-of-Distribution Detection in Skin Images}
Due to the high-level of inter-class similarity and intra-class variation in dermoscopic image classification~\cite{liu2020semi}, OOD detection for skin images has also been studied. For example, Li et al. \cite{li2020out} propose an OOD detection algorithm that fuses deep neural networks and parametric-free isolation forest. Bagchi et al. \cite{bagchi2020learning} use an ensemble model to classify in-distribution data and design a CS-KSU module collection to detect OOD data. Roy et al. \cite{roy2022does} propose a new HOD loss and find that the use of recent representation learning methods and a suitable ensemble strategy can significantly improve performance, and then finally introduce a cost matrix to estimate the downstream clinical impact. Kim et al. \cite{kim2022out} add perturbations to the data to maximize the variance of the OOD samples and apply subset scanning in the latent space representation. Mohseni et al. \cite{mohseni2021out} design a network called BinaryHeads capable of simultaneously classifying ID/OOD data. Unlike existing work, we propose a multi-scale detection framework that integrates one-class SVM and adapted Gram matrix for OOD detection in skin images.

\section{Method}
\begin{figure*}[htbp]
\setlength{\belowcaptionskip}{-0.6cm}
    \centering
    \includegraphics[width=0.95\textwidth]{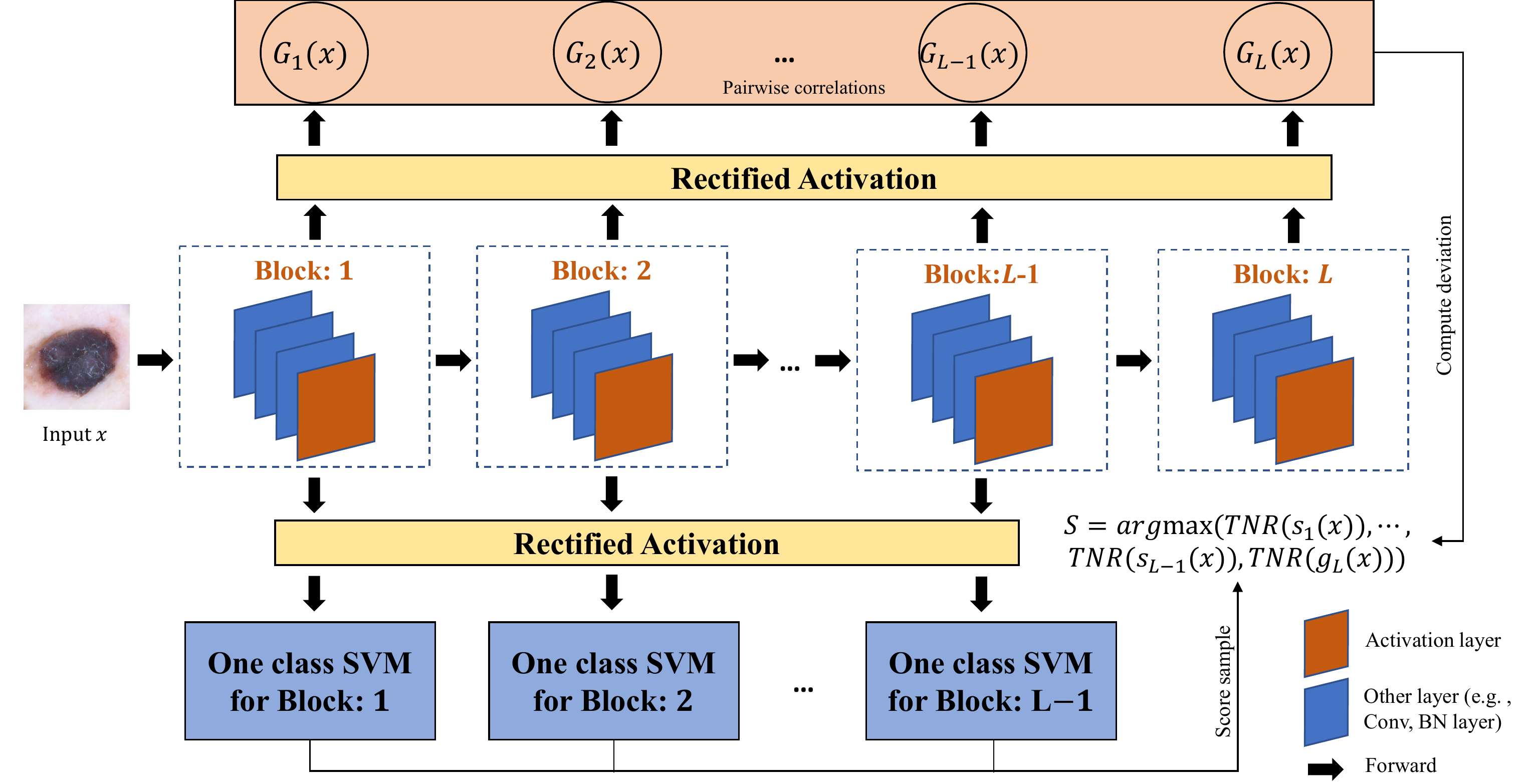}
    \caption{The overview framework of the proposed multi-scale OOD detection framework.}
    \label{fig:framework}
\end{figure*}

In this section, we first introduce a rectified one-class support vector machine to detect OOD data from the output of early network layers, and then use the adapted Gram matrix to detect OOD data from the penultimate network layer. Additionally, we adopt a multi-scale detection framework to integrate the above techniques and further improve the ability of the neural network for detecting out-of-distribution samples. See Figure~\ref{fig:framework} for an overview of the proposed method.

\subsection{Rectified One-Class Support Vector Machine}
Inspired by \cite{abdelzad2019detecting}, we also regard the OOD detection of skin images as an one-class classification problem, and use an one-class Support Vector Machine (SVM) to analyze the shallow features of the images after passing them through the neural network. One-class SVM aims to find a hyperplane in the vector space, so that the hyperplane is the farthest from the zero point, and all normal data (in-distribution data) are on the same side of the hyperplane. Therefore, one-class SVM can produce good results in the detection of out-of-distribution data. In addition, because the high-dimensional features generated by deep neural networks will have a negative impact on one-class SVM, we reduce its dimension in the width and height dimensions, following~\cite{abdelzad2019detecting}: 
\begin{equation}
    f_{k}^l(x) = \frac{1}{w * h} \sum_{i=1}^w \sum_{j=1}^h |f_{ijk}^l|
\end{equation}
where $f_{k}^l(x)$ denotes the $k$-th feature map at the $l$-th layer given input $x$. $w$, $h$ are the width and height of the feature map, $f_{ijk}^l$ is the $(i, j, k)$-th element of $f^l(x)$.

Since the main purpose of out-of-distribution detection is to make the model only confident in the samples within the distribution, we introduce rectified activation~\cite{sun2021react} in order to further improve the detection accuracy and to reduce the frequency of high confidence predictions of out-of-distribution data. By setting a threshold $c$ in the feature extraction layer, the output of layer $f(x)$ is compared with $c$ : $V = min(f(x), c)$, where $V$ is the value to be fed into the classifier. We choose the activation layers of the backbone network as the feature extraction layer, so that the features can be resued for computing the adapted Gram matrix, as outlined in the next subsection. Specifically, we add a rectified activation operation to the output of each layer to make the activation pattern closer to the actual distribution, and then send it to the one-class SVM at the corresponding layer.

\subsection{Adapted Gram Matrix}
Gram matrix is proposed by \cite{sastry2019detecting} and used for out-of-distribution detection, and the adapted Gram matrix is improved by~\cite{pacheco2020out} for the skin cancer classification task. Here, we begin by revisiting the procedure for computing the Gram matrix. First, we obtain the relationship between the features of layer $l$ through a $p$-order matrix, which is written as: $G_{l}^p = (r_{l}^p {r_{l}^p}^T)$. Here, $G$ is the Gram matrix, $r$ is the representation of the $l$-th layer, and additional regularization is performed to make all the values of $G$ in $(0, 1)$, so as to ensure that each maximum and minimum value is calculated from the same interval. At the same time, in order to reduce the complexity of the algorithm, the above features are obtained only in the activation layers of the network, and $p$ = 1 (effective for skin disease classification). Next, in the training set, a maximum and a minimum value are determined according to $G$: 
\begin{equation}
\begin{aligned}
    \lambda_{l}^p = min[G_{l}^p(f(x))] \\
    \Lambda_{l}^p = max[G_{l}^p(f(x))]
\end{aligned}
\end{equation}
where $f(\cdot)$ represents the network, and $x$ is the input image. In practice, we use the row-wise sum of $G_l^p$ for computing the maximum and minimum values, as proposed in~\cite{pacheco2020out}.

Finally, when an unknown image inputs, the distribution deviation between the unknown image and data in the distribution (i.e., the training set) can be calculated with normalization using the previously saved maximum and minimum values:
\begin{equation}
    \delta(\lambda, \Lambda, G) = \begin{cases}
        0, & \lambda \leq G \leq \Lambda \\
        \frac{\lambda-G}{|\lambda|}, & G < \lambda \\
        \frac{G-\Lambda}{|\Lambda|}, & G > \Lambda \\
    \end{cases}
\end{equation}
\begin{equation}
    g(x) = \sum_{l=1}^L \sum_{k=1}^K \frac{\delta(\lambda_{l}^p[k], \Lambda_{l}^p[k], G_{l}^p(f(x))[k])}{E_{Va}[\delta_{l}]}
\end{equation}
where $g(\cdot)$ denotes the output of the adapted Gram matrix in different layers, $k$ denotes the $k$-th feature map of a certain layer, $E_{Va}[\delta_{l}]$ denotes the expected deviation at layer $l$ computed on the validation set. The deviation value can be compared with the threshold determined by the percentile of the total deviation to judge whether a sample is ID or OOD. We also introduce the rectified activation in penultimate layer which can make the output of the activation layer closer to the true distribution, and also make the total deviation of the adapted Gram matrix closer to the well-behaved case.

\subsection{Multi-Scale Detection Framework}
The design of a multi-scale network is widely used in the field of object detection~\cite{zhang2021multi,liu2021samnet}, among other computer vision tasks. For large objects, its semantic information will appear in the feature maps at deeper layers; for small objects, the semantic information appears in the feature maps at shallow layers. Inspired by this, we design a multi-scale detection framework using different OOD detection approaches mentioned above at different network layers to further improve the ability of our model for recognizing out-of-distribution samples. In the early layers, rectified one-class support vector machine is used for out-of-distribution detection. When the image is in the penultimate layer of the network, we use the adapted Gram matrix. In our multi-scale detection framework, the final ID/OOD classification result of an image $x$ depends on the performance of the two methods above on different out-of-distribution datasets. We choose our final normality score computing layer $S$ by the maximum TNR (True Negative Rate) of the one-class SVM and the adapted Gram matrix at every layer:
\begin{equation}
    \text{TNR} = \frac{\text{TN}}{\text{TN} + \text{FP}}
\end{equation}
\begin{equation}
\label{eq:final}
    S = \operatorname*{arg\,max}_{s_1,...,s_{L-1},g_L}[TNR(s_{1}(x)), \cdots, TNR(s_{L-2}(x)), TNR(s_{L-1}(x)), TNR(g_{L}(x))]
\end{equation}
where TN is the number of true negatives and FP is the number of false positives, $s_{l}(\cdot)$ denotes output of the one-class SVM at the $l$-th layer and $g_{L}(\cdot)$ denotes the output of the adapted Gram matrix at the $L$-th layer. It should be noted that our method requires an OOD dataset available at training in order to compute TNR. When OOD data are not available, we could use empirical values of $S$ for a specific model such as those listed in Table \ref{tab:layer}. The final ID/OOD classification result will be produced by comparing the output with a threshold $\theta$:
\begin{equation}
    D(x) = \begin{cases}
            1, & S(x) > \theta \\
            0, & S(x) < \theta
            \end{cases}
\end{equation}
where $D(\cdot)$ represents the final ID/OOD classification result. If the normality score of an input $x$ is greater than $\theta$, it belongs to ID. It should be noted that the results of the same out-of-distribution dataset are generated by the same chosen network layer. In the experiments that follow, $\theta$ is set to 0.95.

\section{Experiments}
\subsection{Setup}
\subsubsection{ID/OOD Datasets} 
\begin{figure*}[ht!]
\setlength{\belowcaptionskip}{-0.2cm}
    \centering
    \subfigure[]{
    \includegraphics[width=0.12\textwidth]{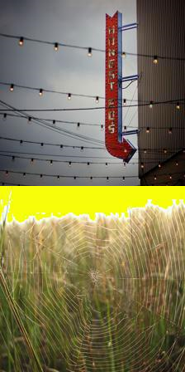}
    }
    \quad
    \subfigure[]{
    \includegraphics[width=0.12\textwidth]{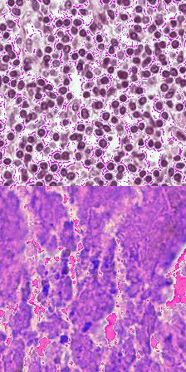}
    }
    \quad
    \subfigure[]{
    \includegraphics[width=0.12\textwidth]{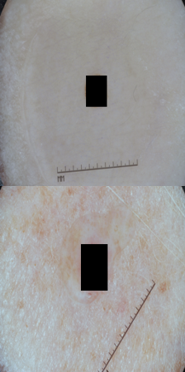}
    }
    \quad
    \subfigure[]{
    \includegraphics[width=0.12\textwidth]{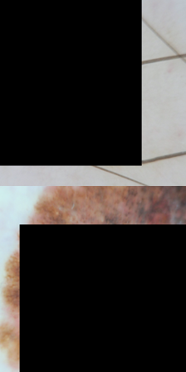}
    }
    \quad
    \subfigure[]{
    \includegraphics[width=0.12\textwidth]{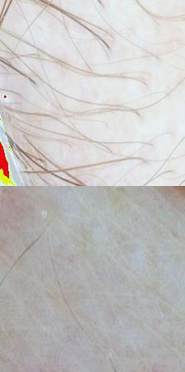}
    }
    \quad
    \subfigure[]{
    \includegraphics[width=0.12\textwidth]{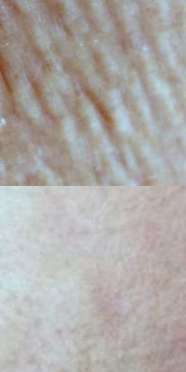}
    }
    \caption{Samples of six non-overlapping datasets as OOD data for detection. From (a) to (f): Imagenet, NCT, BBOX, BBOX70, Derm-skin, Clinical.}
    \label{fig:sample}
\end{figure*}

In our experiments, ISIC 2019~\cite{codella2018skin,combalia2019bcn20000,tschandl2018ham10000} is regarded as the ID dataset, which consists of eight different categories of skin diseases, with a total of 25331 images. The division and preprocessing of the dataset are consistent with~\cite{pacheco2020out}. In addition, we use the following six non-overlapping OOD datasets for detection, with example images shown in Figure~\ref{fig:sample}:
\vspace{-2mm}
\paragraph{ImageNet.} Contains 3000 images randomly sampled from ImageNet \cite{deng2009large} test set.
\vspace{-2mm}
\paragraph{NCT.} There are 9 classes in the original NCT-CRC-HE-7K \cite{kather2019predicting} dataset. The data used for detection are 150 randomly sampled human colorectal cancer (CRC) images from each class.
\vspace{-2mm}
\paragraph{BBOX.} It contains 2025 skin disease images with bounding boxes after successful segmentation by U-net trained on the ISIC 2017 segmentation dataset. \cite{ronneberger2015u}.
\vspace{-2mm}
\paragraph{BBOX70.} The acquisition method of this dataset is the same as that of BBOX, but the bounding box will cover 70 \% of the whole skin lesion.
\vspace{-2mm}
\paragraph{Derm-skin.} It contains 1565 images selected from ISIC 2019. These images belong to the area that does not contain lesions after cropping.
\vspace{-2mm}
\paragraph{Clinical.} Contains 723 healthy skin images collected from social networks.

\subsubsection{Pre-trained Models and Parameters}
For a fair comparison, we directly use the model pre-trained on the ISIC 2019 training set used by~\cite{pacheco2020out} as our backbone network, including DenseNet-121 \cite{huang2017densely}, MobileNet-v2 \cite{sandler2018mobilenetv2}, ResNet-50 \cite{he2016deep} and VGGNet-16 \cite{simonyan2014very}. All models are optimized using the Adam algorithm. The initial learning rate is set to 0.0001 and the batch size is set to 40. Among them, the learning rate is decreased by a factor of 0.2 after the model failed to optimize the validation loss for 15 consecutive epochs. The balanced accuracy of the four pre-trained models on the ISIC 2019 test set are 82.3\%, 81.2\%, 82\% and 82.5\% respectively.

As for the one-class SVM, we use RBF kernel for training and set $\nu$ to 0.001, which is consistent with the experimental setting of \cite{abdelzad2019detecting}. We select the output layer with the highest TNR as the network layer used to calculate the normality score. The output layers corresponding to each out-of-distribution dataset and the backbone network are shown in Table \ref{tab:layer}. Under four different models, the rectified activation operation thresholds used for the adapted Gram matrix are 1.0, 0.8, 0.6 and 0.7, while the rectified activation operation thresholds for one-class SVM are all set to 1.0.

\begin{table}[ht!]
    \centering
    \caption{Layer selected for calculating the normality score corresponding to each out-of-distribution dataset and network backbone.}
    \label{tab:layer}
    \setlength{\tabcolsep}{6pt}    
    \renewcommand{\arraystretch}{1.2}
    \begin{tabular}{l|cccccc}
    \hline
    \multicolumn{1}{c|}{\multirow{2}{*}{\textbf{Model}}} & \multicolumn{6}{c}{\textbf{Selected Layer}}                                                                                                                                                 \\ \cline{2-7} 
    \multicolumn{1}{c|}{}                       & \multicolumn{1}{l|}{Imagenet} & \multicolumn{1}{l|}{NCT} & \multicolumn{1}{l|}{BBOX} & \multicolumn{1}{l|}{BBOX70} & \multicolumn{1}{l|}{Derm-skin} & \multicolumn{1}{l}{Clinical} \\ \hline
    DenseNet-121                                 & \multicolumn{1}{c|}{25}       & \multicolumn{1}{c|}{39}  & \multicolumn{1}{c|}{13}   & \multicolumn{1}{c|}{9}      & \multicolumn{1}{c|}{3}         & 74                            \\
    MobileNet-v2                                 & \multicolumn{1}{c|}{18}       & \multicolumn{1}{c|}{25}  & \multicolumn{1}{c|}{19}   & \multicolumn{1}{c|}{10}     & \multicolumn{1}{c|}{2}         & 24                            \\
    ResNet-50                                    & \multicolumn{1}{c|}{13}       & \multicolumn{1}{c|}{8}   & \multicolumn{1}{c|}{16}    & \multicolumn{1}{c|}{5}      & \multicolumn{1}{c|}{1}         & 10                            \\
    VGGNet-16                                    & \multicolumn{1}{c|}{8}        & \multicolumn{1}{c|}{8}   & \multicolumn{1}{c|}{9}    & \multicolumn{1}{c|}{1}      & \multicolumn{1}{c|}{2}         & 13                             \\ \hline
    \end{tabular}
\end{table}

\begin{table}[ht!]
\centering
\caption{Comparison with state-of-the-art methods in out-of-distribution detection using different backbone networks and datasets.}
\label{tab:sota}
\setlength{\tabcolsep}{6pt}    
\renewcommand{\arraystretch}{1.8}
\large
\resizebox{\textwidth}{!}{%
\begin{tabular}{c|c|ccc}
\hline
\multirow{2}{*}{\textbf{Model}}                                                   & \multirow{2}{*}{\textbf{OOD set}} & \multicolumn{1}{c|}{\textbf{AUROC}}                                   & \multicolumn{1}{c|}{\textbf{Detection Accuracy}}                      & \textbf{TNR @ TPR 95\%}                          \\ \cline{3-5} 
                                                                         &                          & \multicolumn{3}{c}{Baseline/\underline{ODIN}/\underline{Mahalanobis}/Gram-OOD/Gram-OOD* (w/rectified activation)/Ours}                                                                                                \\ \hline
\multirow{6}{*}{\begin{tabular}[c]{@{}c@{}}DenseNet\\ -121\end{tabular}} & ImageNet                 & \multicolumn{1}{c|}{59.1 / \underline{83.8} / \underline{99.9} / 97.0 / 97.3 / \textbf{99.3}} & \multicolumn{1}{c|}{56.6 / \underline{78.1} / \underline{99.1} / 92.0 / 93.0 / \textbf{96.6}} & 9.30 / \underline{50.0} / \underline{99.9} / 80.7 / 86.0 / \textbf{97.7} \\
                                                                         & NCT                      & \multicolumn{1}{c|}{36.7 / \underline{82.0} / \underline{98.9} / 99.4 / 99.4 / \textbf{100.}} & \multicolumn{1}{c|}{50.1 / \underline{75.0} / \underline{98.7} / 97.1 / 98.1 / \textbf{99.7}} & 1.44 / \underline{32.5} / \underline{98.7} / 98.9 / 99.9 / \textbf{100.} \\
                                                                         & BBOX                     & \multicolumn{1}{c|}{77.3 / \underline{90.6} / \underline{98.3} / 98.1 / 97.5 / \textbf{99.4}} & \multicolumn{1}{c|}{69.8 / \underline{83.7} / \underline{95.3} / 94.5 / 93.6 / \textbf{97.3}} & 27.9 / \underline{68.8} / \underline{94.8} / 88.0 / 88.2 / \textbf{98.4} \\
                                                                         & BBOX70                   & \multicolumn{1}{c|}{89.4 / \underline{99.8} / \underline{100.} / 99.7 / 99.8 / \textbf{100.}} & \multicolumn{1}{c|}{84.9 / \underline{98.1} / \underline{99.9} / 99.0 / 99.2 / \textbf{99.9}} & 36.6 / \underline{99.3} / \underline{100.} / 99.9 / 100. / \textbf{100.} \\
                                                                         & Derm-skin                & \multicolumn{1}{c|}{74.4 / \underline{86.8} / \underline{96.2} / 96.5 / \textbf{96.6} / 96.4} & \multicolumn{1}{c|}{67.3 / \underline{78.3} / \underline{89.7} / 90.9 / 91.0 / \textbf{91.4}} & 22.8 / \underline{46.2} / \underline{81.4} / 78.0 / 81.8 / \textbf{87.5} \\
                                                                         & Clinical                 & \multicolumn{1}{c|}{72.5 / \underline{69.5} / \underline{96.1} / 96.6 / 96.3 / \textbf{98.1}} & \multicolumn{1}{c|}{67.3 / \underline{65.8} / \underline{90.1} / 91.1 / 91.1 / \textbf{92.9}} & 18.5 / \underline{25.2} / \underline{81.7} / 82.8 / 84.6 / \textbf{90.7} \\ \hline
\multirow{6}{*}{\begin{tabular}[c]{@{}c@{}}MobileNet\\ -v2\end{tabular}} & ImageNet                 & \multicolumn{1}{c|}{61.9 / \underline{86.8} / \underline{99.7} / 97.2 / 98.4 / \textbf{99.6}} & \multicolumn{1}{c|}{58.5 / \underline{81.8} / \underline{98.5} / 92.1 / 94.5 / \textbf{97.3}} & 12.4 / \underline{36.6} / \underline{99.8} / 84.3 / 92.6 / \textbf{98.8} \\
                                                                         & NCT                      & \multicolumn{1}{c|}{75.7 / \underline{72.2} / \underline{99.9} / 99.4 / 99.7 / \textbf{100.}} & \multicolumn{1}{c|}{68.2 / \underline{69.9} / \underline{99.3} / 97.4 / 98.9 / \textbf{99.4}} & 25.4 / \underline{33.3} / \underline{100.} / 99.3 / 100. / \textbf{100.} \\
                                                                         & BBOX                     & \multicolumn{1}{c|}{56.3 / \underline{95.3} / \underline{99.3} / 97.3 / 98.8 / \textbf{99.8}} & \multicolumn{1}{c|}{56.2 / \underline{90.0} / \underline{95.6} / 94.4 / 97.0 / \textbf{98.5}} & 6.70 / \underline{71.9} / \underline{96.3} / 86.9 / 98.6 / \textbf{100.}  \\
                                                                         & BBOX70                   & \multicolumn{1}{c|}{72.6 / \underline{97.9} / \underline{99.9} / 99.8 / 99.9 / \textbf{100.}} & \multicolumn{1}{c|}{68.1 / \underline{96.0} / \underline{99.8} / 99.0 / 99.5 / \textbf{99.9}} & 13.4 / \underline{92.9} / \underline{100.} / 100. / 100. / \textbf{100.} \\
                                                                         & Derm-skin                & \multicolumn{1}{c|}{65.1 / \underline{79.4} / \underline{92.6} / 94.2 / \textbf{94.7} / 93.5} & \multicolumn{1}{c|}{59.8 / \underline{71.8} / \underline{86.1} / 87.1 / \textbf{88.4} / 87.3} & 18.8 / \underline{40.8} / \underline{64.2} / 66.7 / 75.2 / \textbf{79.4} \\
                                                                         & Clinical                 & \multicolumn{1}{c|}{62.9 / \underline{78.3} / \underline{97.6} / 95.3 / 96.3 / \textbf{96.8}} & \multicolumn{1}{c|}{59.6 / \underline{71.7} / \underline{92.6} / 89.6 / \textbf{90.7} / 89.8} & 14.2 / \underline{27.8} / \underline{85.5} / 77.9 / 83.5 / \textbf{84.4} \\ \hline
\multirow{6}{*}{\begin{tabular}[c]{@{}c@{}}ResNet\\ -50\end{tabular}}    & ImageNet                 & \multicolumn{1}{c|}{60.1 / \underline{83.9} / \underline{99.9} / 97.9 / 97.9 / \textbf{99.5}} & \multicolumn{1}{c|}{57.6 / \underline{77.0} / \underline{99.2} / 92.9 / 93.2 / \textbf{96.9}} & 8.50 / \underline{49.2} / \underline{99.9} / 86.6 / 87.4 / \textbf{98.2} \\
                                                                         & NCT                      & \multicolumn{1}{c|}{67.4 / \underline{93.3} / \underline{99.9} / 99.8 / 99.9 / \textbf{100.}} & \multicolumn{1}{c|}{64.6 / \underline{86.0} / \underline{99.6} / 98.4 / 99.0 / \textbf{99.9}} & 8.40 / \underline{70.2} / \underline{100.} / 99.9 / 100. / \textbf{100.} \\
                                                                         & BBOX                     & \multicolumn{1}{c|}{69.7 / \underline{74.5} / \underline{99.8} / 97.9 / 99.4 / \textbf{99.6}} & \multicolumn{1}{c|}{65.1 / \underline{69.7} / \underline{98.0} / 94.2 / \textbf{97.5} / 97.4} & 11.7 / \underline{34.9} / \underline{99.6} / 88.4 / \textbf{99.2} / 98.7 \\
                                                                         & BBOX70                   & \multicolumn{1}{c|}{71.6 / \underline{99.7} / \underline{99.9} / 99.9 / 100. / \textbf{100.}} & \multicolumn{1}{c|}{72.2 / \underline{97.9} / \underline{99.9} / 99.5 / 99.7 / \textbf{99.8}} & 8.90 / \underline{99.2} / \underline{100.} / 100. / 100. / \textbf{100.} \\
                                                                         & Derm-skin                & \multicolumn{1}{c|}{72.1 / \underline{87.2} / \underline{96.0} / \textbf{96.1} / 95.1 / 95.0} & \multicolumn{1}{c|}{66.8 / \underline{80.2} / \underline{89.7} / \textbf{90.1} / 88.5 / 89.5} & 14.8 / \underline{57.9} / \underline{81.1} / 74.8 / 73.9 / \textbf{83.8} \\
                                                                         & Clinical                 & \multicolumn{1}{c|}{62.0 / \underline{71.4} / \underline{95.1} / 97.2 / \textbf{97.4} / 96.9} & \multicolumn{1}{c|}{59.7 / \underline{67.0} / \underline{88.9} / 91.2 / \textbf{91.8} / 90.7} & 8.30 / \underline{23.6} / \underline{73.4} / 84.7 / 85.9 / \textbf{86.4} \\ \hline
\multirow{6}{*}{\begin{tabular}[c]{@{}c@{}}VGGNet\\ -16\end{tabular}}    & ImageNet                 & \multicolumn{1}{c|}{46.6 / \underline{82.9} / \underline{99.4} / 96.3 / 95.3 / \textbf{98.5}} & \multicolumn{1}{c|}{50.6 / \underline{79.0} / \underline{98.0} / 90.2 / 90.5 / \textbf{94.2}} & 5.90 / \underline{25.1} / \underline{99.3} / 77.6 / 81.3 / \textbf{93.3} \\
                                                                         & NCT                      & \multicolumn{1}{c|}{57.4 / \underline{72.1} / \underline{99.2} / 99.6 / 99.8 / \textbf{100.}} & \multicolumn{1}{c|}{55.5 / \underline{69.5} / \underline{98.9} / 97.9 / 99.0 / \textbf{99.4}} & 10.7 / \underline{16.6} / \underline{99.2} / 99.7 / 100. / \textbf{100.} \\
                                                                         & BBOX                     & \multicolumn{1}{c|}{74.9 / \underline{86.9} / \underline{99.9} / 97.9 / 98.4 / \textbf{99.0}} & \multicolumn{1}{c|}{67.4 / \underline{81.3} / \underline{98.6} / 94.0 / 95.1 / \textbf{95.5}} & 30.3 / \underline{64.6} / \underline{99.8} / 86.5 / 94.0 / \textbf{95.7} \\
                                                                         & BBOX70                   & \multicolumn{1}{c|}{81.7 / \underline{99.9} / \underline{100.} / 99.9 / 100. / \textbf{100.}} & \multicolumn{1}{c|}{83.1 / \underline{99.2} / \underline{99.9} / 99.7 / 99.7 / \textbf{99.8}} & 5.40 / \underline{99.0} / \underline{100.} / 100. / 100. / \textbf{100.} \\
                                                                         & Derm-skin                & \multicolumn{1}{c|}{67.1 / \underline{93.1} / \underline{91.4} / 96.0 / 93.6 / \textbf{96.1}} & \multicolumn{1}{c|}{61.4 / \underline{87.1} / \underline{83.6} / 89.8 / 89.4 / \textbf{91.9}} & 21.1 / \underline{78.6} / \underline{65.8} / 79.8 / 80.6 / \textbf{88.6} \\
                                                                         & Clinical                 & \multicolumn{1}{c|}{66.3 / \underline{72.4} / \underline{97.2} / \textbf{95.7} / 93.8 / 95.4} & \multicolumn{1}{c|}{62.1 / \underline{68.3} / \underline{91.6} / \textbf{89.8} / 89.1 / 87.3} & 15.0 / \underline{31.3} / \underline{84.3} / 80.7 / \textbf{81.1} / 80.9 \\ \hline
\end{tabular}%
}
\end{table}

\subsubsection{Evaluation Metrics}
We adopt three evaluation metrics commonly used in out-of-distributin detection: area under the ROC curve (AUROC)~\cite{sastry2019detecting}; the maximum achievable classification accuracy across all possible thresholds in distinguishing among in-distribution and out-of-distribution samples (Detection Accuracy)~\cite{sastry2019detecting}; true negative rate when the true positive rate is as high as 95\% (TNR @ TPR 95\%), where TNR can be computed as TN/(TN $+$ FP), where TN and FP represent true negatives and false positives~\cite{sastry2019detecting}.

\begin{figure*}[ht!]
    \centering
    \includegraphics[width=0.92\textwidth]{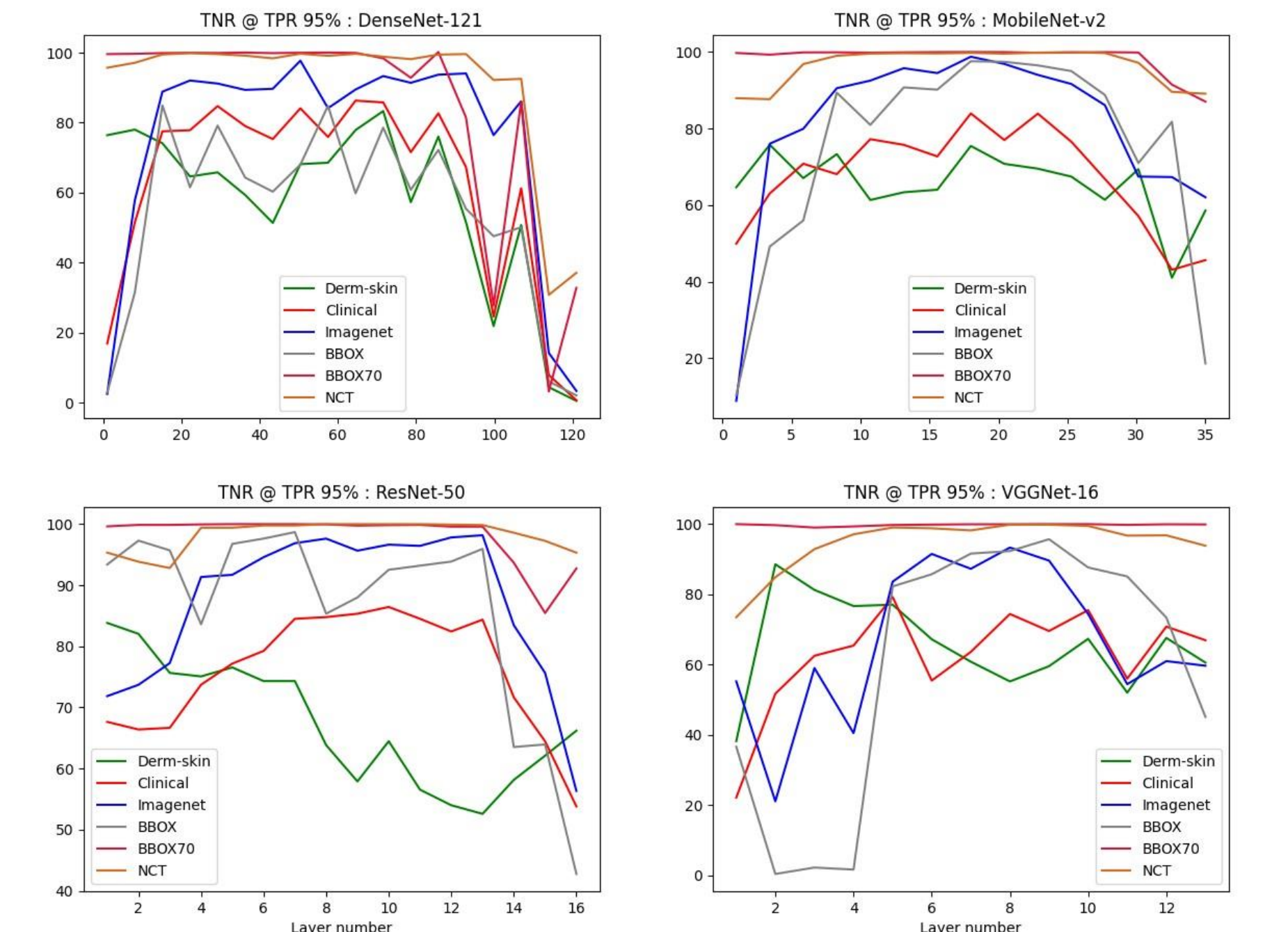}
    \caption{Influence of layer selected to compute the normality score on TNR @ TPR 95\% with different models after smoothing.}
    \label{fig:layer}
\end{figure*}

\begin{figure*}[ht!]
    \centering
    \includegraphics[width=0.92\textwidth]{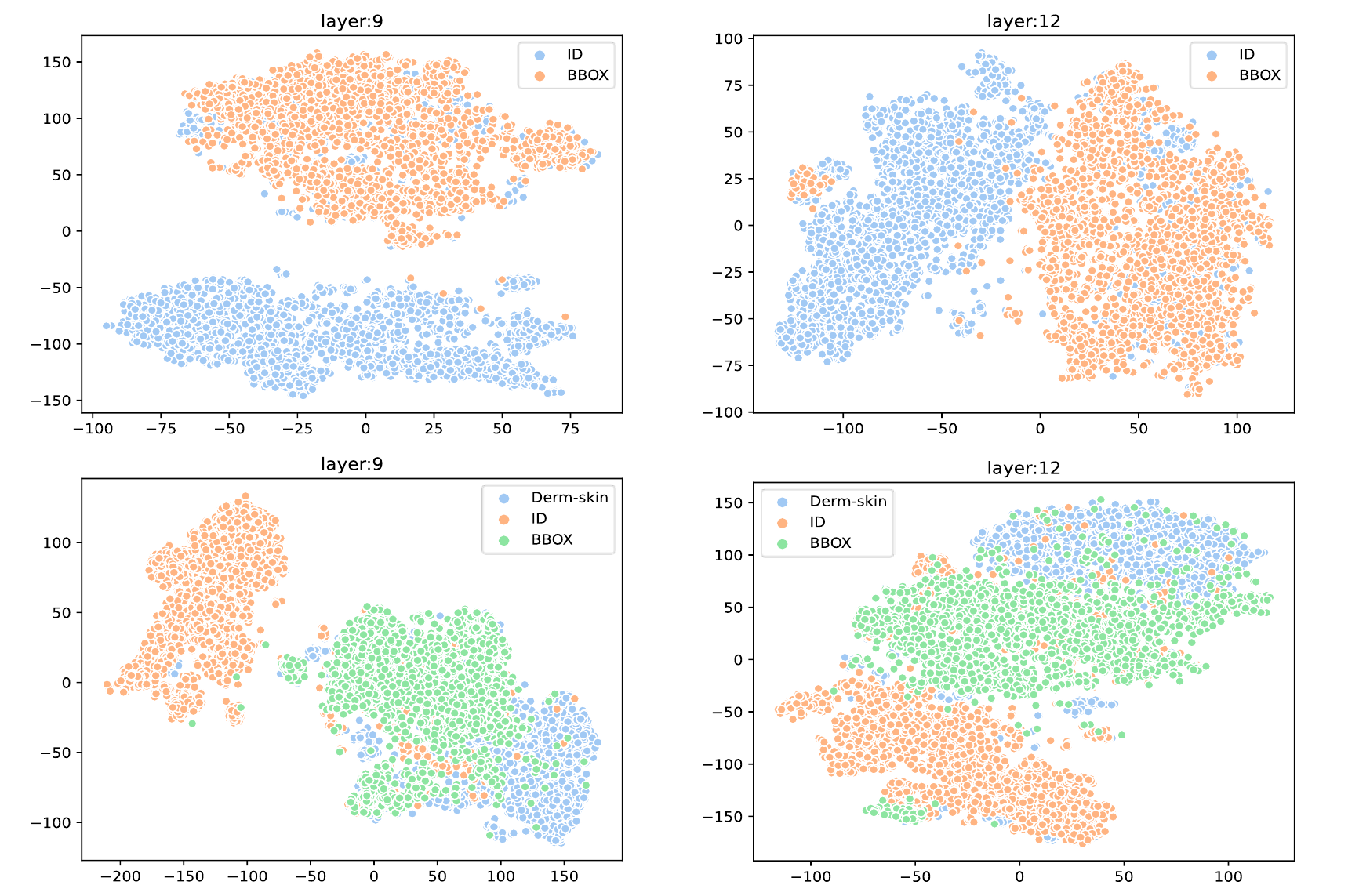}
    \caption{t-SNE~\cite{van2008visualizing} visualization of features extracted from the layer selected for computing the normality score (layer 9) and the penultimate layer~(layer 12).}
    \label{fig:svmmlayer}
\end{figure*}

\subsection{Results}
\subsubsection{Comparison with State-of-the-Art OOD Algorithms} In order to better verify the effectiveness of our proposed multi-scale detection framework, we compare the framework with the following recently published methods: ODIN~\cite{liang2017enhancing}, Mahalanobis~\cite{lee2018simple}, Gram-OOD~\cite{sastry2019detecting} and Gram-OOD* \cite{pacheco2020out}. It should be noted that ODIN and Mahalanobis have been fine-tuned on the OOD dataset. Also, our method requires OOD data during training in order to choose the best-performing layer for OOD classification. See Figure~\ref{fig:layer} for a detailed performance analysis when we choose different layers. In particular, Gram-OOD* with rectified activations is a special case in our method (i.e., choosing the penultimate layer with the adapted Gram matrix) that still offers a competitive performance.

As shown in Table \ref{tab:sota}, the performance of our framework exceeds that of other recently published methods under most settings. For the evaluation metric TNR @ TPR 95\%, our framework has an obvious advantage, being 5.6\%, 2.1\%, 3.5\%, and 3.6\% higher than Gram-OOD* with rectified activation on average using different models, showing that our framework can better distinguish in distribution~(ID) data and out-of-distribution~(OOD) data based on the shallow neural network layers. The results of ODIN and Baseline show that the traditional method of using the Softmax function for out-of-distribution detection is not suitable for difficult classification tasks such as skin disease anomaly detection.

\subsubsection{Impact of Selecting Different Layers} In Figure \ref{fig:layer}, we present the results of out-of-distribution detection based on the features extracted from different network layers. On the whole, BBOX70 and NCT datasets perform more stably when we extract features from different network layers. Only in the deeper networks will there be the same performance degradation problem as in other datasets, and the possible reason is that it is easier to classify ID/OOD data on these datasets, and the obvious embeddings may be affected as the network grows deeper. On the other four datasets, there are intense performance fluctuations when we extract features from different layers, proving the importance of choosing an appropriate layer for feature extraction. Furthermore, even if the adapted Gram matrix offers a competitive performance as already shown in Table~\ref{tab:sota} (see Gram-OOD* with rectified activations), in most cases it does not provide superior performance with the datasets and models we tested. Besides, we visualize a few examples of the feature embeddings extracted from different layers of the neural network. As Figure~\ref{fig:svmmlayer} shows, the layer for computing the normality score selected by Equation \ref{eq:final} separates ID/OOD data points with a wider margin when compared to the penultimate layer.

\section{Conclusion}
In this paper, to enhance the security of skin disease identification systems, we investigate the classification-based out-of-distribution detection problem in dermoscopic images. Specifically, we propose a multi-scale detection framework that uses different classifiers at the various layers of the neural network to detect and compare feature embeddings, and select the layer with the best performance to compute a normality score. In addition, we adopt a rectified activation operation before feeding features into classifiers to ensure that the features are closer to a well-behaved distribution. We hope that our modest attempt could provide some useful insights for future research on out-of-distribution detection in skin images and beyond.

\vspace{5mm}
\footnotesize{\noindent \textbf{Acknowledgments.} This work is supported by NSFC (61703195), Fujian NSF (2022J011112, 2020J01828), Guangdong NSF (2019A1515011045), Science and Technology Program of Guangzhou (202102020692), the Open Program of The Key Laboratory of Cognitive Computing and Intelligent Information Processing of Fujian Education Institutions, Wuyi University (KLCCIIP2020202), the Open Fund of Fujian Provincial Key Laboratory of Information Processing and Intelligent Control, Minjiang University (MJUKF-IPIC202102), the CAAI-Huawei MindSpore Open Fund, and Fuzhou Technology Planning Program (2021-ZD-284).}


{\small
\bibliographystyle{ieee}
\bibliography{reference}
}

\end{document}